\documentclass[runningheads]{llncs}

\usepackage{eccv}

\usepackage{eccvabbrv}

\usepackage{graphicx}
\usepackage{booktabs}

\usepackage[accsupp]{axessibility}  

\usepackage{hyperref}

\usepackage{orcidlink}

\begin{document}

\title{Unsupervised Tomato Split Anomaly Detection using Hyperspectral Imaging and Variational Autoencoders} 

\titlerunning{Unsupervised Tomato Split Anomaly Detection using HSI and VAE}

\author{Mahmoud Abdulsalam\inst{1} \and
Usman Zahidi\inst{1} \and
Bradley Hurst\inst{1} \and 
Simon Pearson \inst{1}\and Grzegorz Cielniak \inst{1} \and James Brown \inst{1}}

\authorrunning{M.~Abdulsalam et al.}

\institute{University of Lincoln, Brayford Pool, Lincoln, LN6 7TS, UK \\
\email{\{MAbdulsalam, UZahidi, BrHurst, spearson, gcielniak, JamesBrown\}@lincoln.ac.uk}}

\maketitle

\begin{abstract}
  Tomato anomalies/damages pose a significant challenge in greenhouse farming. While this method of cultivation benefits from efficient resource utilization, anomalies can significantly degrade the quality of farm produce. A common anomaly associated with tomatoes is splitting, characterized by the development of cracks on the tomato skin, which degrades its quality. Detecting this type of anomaly is challenging due to dynamic variations in appearance and sizes, compounded by dataset scarcity. We address this problem in an unsupervised manner by utilizing a tailored variational autoencoder (VAE) with hyperspectral input. Preliminary analysis of the dataset enabled us to select the optimal range of wavelengths for detecting this anomaly. Our findings indicate that the 530nm - 550nm range is suitable for identifying tomato dry splits. The proposed VAE model achieved a 97\% detection accuracy for tomato split anomalies in the test data. The analysis on reconstruction loss allow us to not only detect the anomalies but also to some degree estimate the anomalous regions.
\end{abstract}

\section{Introduction}
\label{sec:intro}
Tomato fruit is rich in multivitamins, including Vitamin C and E, and contains the antioxidant lycopene. This fruit has been linked to reducing the risk of certain cardiovascular diseases, making it widely consumed \cite{chang2006comparisons}. Tomato cultivation is often conducted in controlled environments, particularly greenhouses, where external environmental factors have minimal impact on the produce \cite{boulard2011environmental}. However, despite these controlled conditions, some tomato anomalies may develop, both within and outside the greenhouse environment, which can degrade the quality of the fruit.

The split anomaly is a common issue in tomato cultivation. It is characterized by cracks on the tomato skin, exposing the internal flesh. This anomaly often varies in appearance and size, with some splits barely visible. Not only does it degrade the quality of the tomato, but it also makes the fruit susceptible to secondary infections. The split anomaly is often caused by a combination of factors, particularly fluctuations in temperature and water supply \cite{RHS}. These factors are difficult to control, given the complexities of the tomato supply chain. Thus, effective detection methods are essential for the early identification of this anomaly. Early detection will increase the quality of the produce, reduce wastage, lower the cost of sorting, and ultimately enhance overall yield.

The use of Hyperspectral Imaging (HSI) allows for the observation of spectral properties of objects by visualizing them through several wavelengths. This capability has driven a shift in agricultural sensing applications from traditional RGB imaging being utilised for detection \cite{adhikari2018tomato, hasan2019deep, salih2020deep, sabrol2016fuzzy, sabrol2016tomato}. In the agricultural domain, hyperspectral imaging is employed for a variety of applications, including weed detection \cite{weed1, pantazi2016active}, pest and disease detection \cite{zhang2019deep, xue2023quantification}, phenotyping \cite{junttila2022close, jarolmasjed2018proximal}, and anomaly detection for infection \cite{li2016fast}, bruises \cite{baranowski2012detection} and cracking \cite{yu2014identification}. 

Anomaly detection through machine learning has been explored from multiple perspectives. A widely adopted approach involves unsupervised anomaly detection techniques, which can be classified into clustering-based, classification-based, reconstruction-based, and self-supervised methods \cite{liu2023self, chandola2009anomaly}. Clustering-based methods learn the distribution of normal data, operating under the premise that anomalous data will occupy low-density regions, distinct from the primary clusters of normal data. Common models for clustering methods include Gaussian Mixture Models (GMM), which represent data distributions as a combination of multiple Gaussian distributions, each with its own mean and variance \cite{zong2018deep}, and the Local Outlier Factor (LOF), which identifies outliers by measuring the local density deviation of a data point relative to its neighbors \cite{breunig2000lof}. One-class SVM (OCSVM) \cite{scholkopf1999support} is a classification based approach that identifies patterns within the data by learning a decision boundary that separates the majority of data points from the rest. The aim is to detect anomalies by classifying new data points as either within the normal region or as outliers based on their deviation from the learned boundary. Self supervised approach leverage the inherent structure of the data to create tasks like predicting missing parts of the data or generating augmentations. The model then learns to distinguish between normal and abnormal patterns by solving these tasks \cite{zhang2022deep,mohseni2020self}. Reconstruction-based models detect anomalies by evaluating reconstruction errors. The principle is that if a model, typically an autoencoder, is trained on normal data, it will struggle to accurately reconstruct anomalous data since it has not encountered such data during training \cite{an2015variational}. A simple but yet an effective methods for anomaly detection. 

Despite these advancements, tomato split anomaly detection remains challenging due to the dynamic appearance and size variations of the anomalies. To address this, we propose an approach that utilises the reconstruction loss of a variational autoencoder  to detect the challenging split anomaly. To the best of our knowledge, there is no available literature that has specifically targeted the tomato split anomaly detection using hyperspectral imaging.

This paper begins with the introduction in Section \ref{sec:intro}. In Section \ref{sec:MM}, we detail the materials and methods, including the data collection, dataset, data pre-processing, and the proposed model. The results are presented and discussed in Section \ref{sec:result}. Finally, we conclude with our findings in Section \ref{sec:conc}.

\section{Materials and Methods}
\label{sec:MM}

\subsection{Data Collection}
\begin{figure}[t!]
  \centering
  \includegraphics[height=5cm]{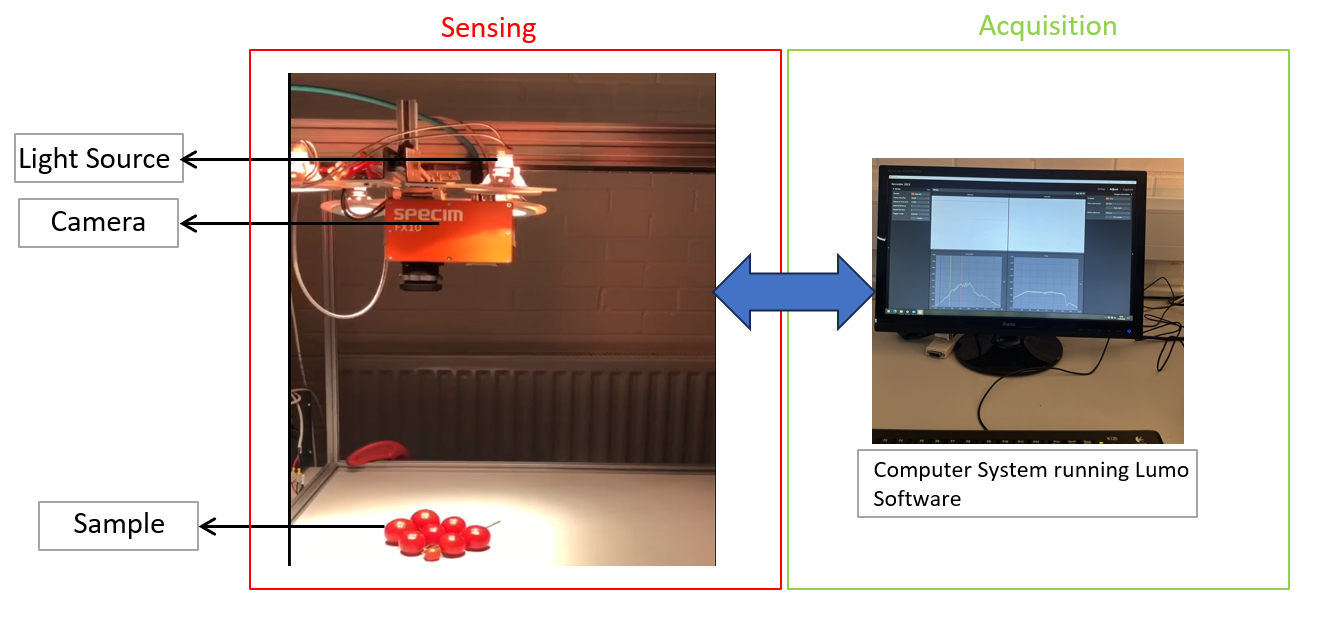}
  \caption{Data collection setup consisting of the lighting source, hyperspectral camera, samples and a computer system.
  }
  \label{fig:setup}
\end{figure}

 The data collection setup includes a camera, lighting source, a computer system running Specim's Lumo software, and the fruit as seen in Fig. \ref{fig:setup}. The camera used is the FX10e hyperspectral camera from Specim \cite{Specim_2024}. This is a line scanner capable of scanning wavelengths ranging from 400nm to 1000nm. For the data collection, we configured the camera to scan 800 lines at a frame rate of 40 frames per second (fps). Each acquired image has dimensions of 1024 (width) × 800 (height) × 448 (depth), where the depth corresponds to the number of wavelengths. The lighting source is attached to the camera rig to ensure sufficient illumination for capturing the spectral properties of the fruit. The Lumo software, equipped with a graphical user interface, facilitates communication with the camera, visualization of the fruit, and the capture/storage of the hyperspectral images. 
\subsection{Dataset}
\begin{figure}[h!]
  \centering
  \includegraphics[height=4.5cm]{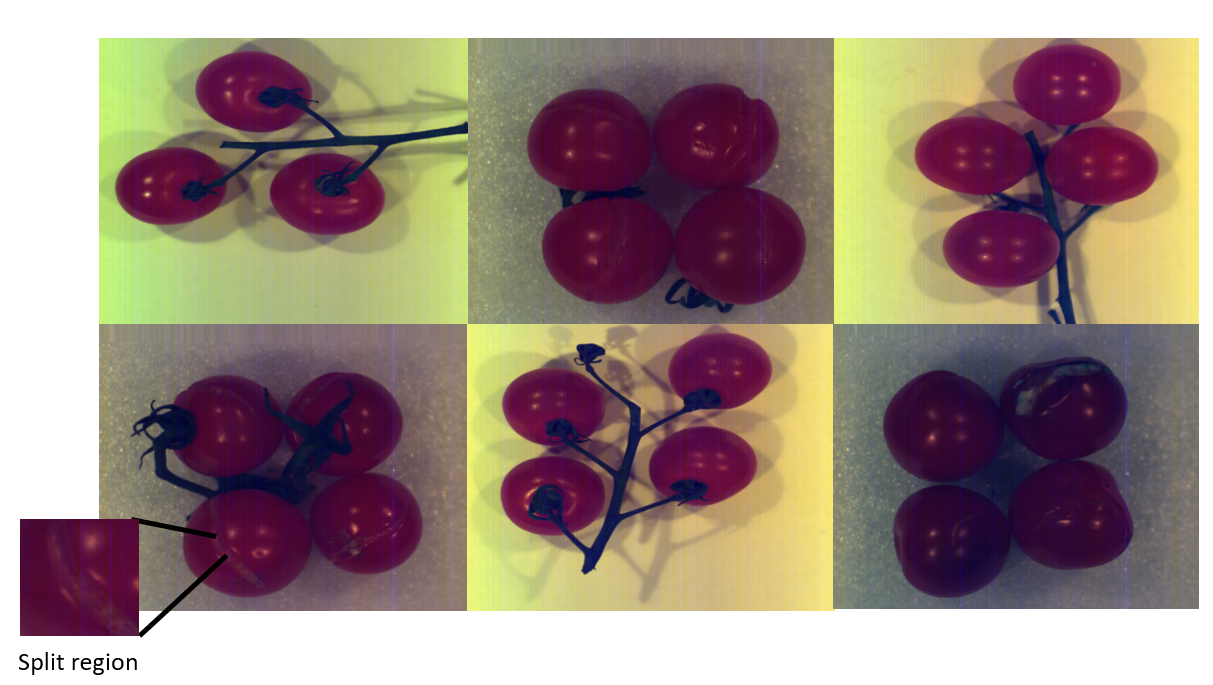}
  \caption{RGB image samples of the dataset for visualisation. The split region is highlighted. 
  }
  \label{fig:dataset_sample}
\end{figure}

A total of 20 scanning sessions of tomatoes yielded 74 tomato samples, which were further spatially augmented (rotation and flipping) to 305 normal instances and 55 anomalous instances. The data collection process was conducted in a dynamic fashion, with tomatoes presented in various orientations to enhance dataset robustness. Each capture included not only the tomatoes but also their stalks and calyxes, ensuring that all relevant parts of the fruit were represented to enrich the dataset. The RGB image samples are shown in Fig. \ref{fig:dataset_sample}.

\subsection{Data Preprocessing}
\label{subsec:process}
 To eliminate sensor noise and calibrate the variation of intensities across wavelengths, the Hyperspectral images  were corrected using the white and dark references according to Equation \ref{eq:calibration}. 

 \begin{equation}
  I_c = \frac{I_i - I_d }{I_w - I_d} 
  \label{eq:calibration}
\end{equation}
where $I_c$ is the calibrated Image, $I_i$ is the original image, $I_d$ is the dark reference image and $I_w$ is the white reference image.

Because the data captured encompasses the tomato bunch, the next pre-processing task involves cropping each individual tomato from the image and masking out the background. For this, we utilized an open source tomato pre-trained YOLOv8 model\cite{JeanN00B,yolov8_ultralytics} to obtain the bounding boxes and masks of the Region of Interest (ROI). The pre-processing pipeline is illustrated in Fig. \ref{fig:yolopreprocess}.
\begin{figure}[t!]
  \centering
  \includegraphics[height=4.5cm]{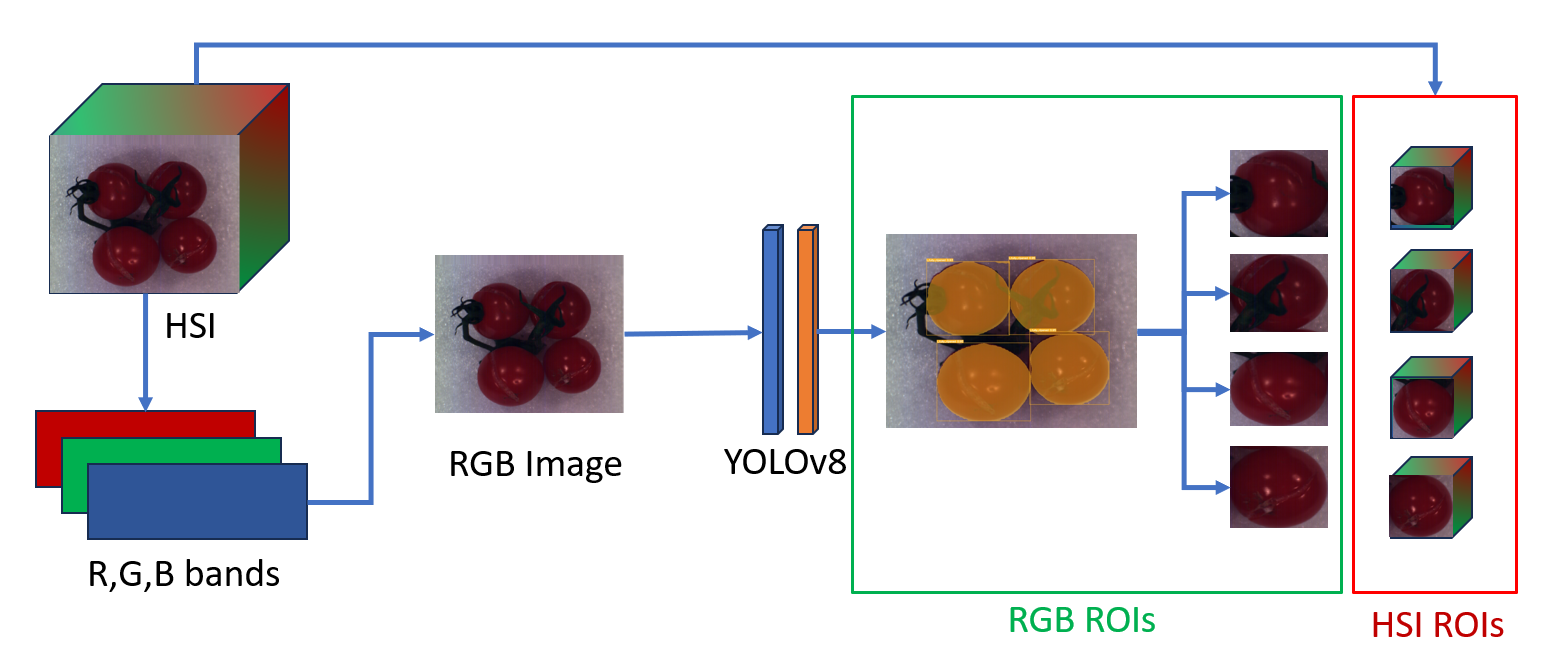}
  \caption{The pre-processing pipeline showing the process of obtaining HSI ROIs (individual tomatoes) from the full HSI (tomato bunch). 
  }
  \label{fig:yolopreprocess}
\end{figure}
Three wavelengths representing the Red, Green, and Blue (RGB) bands were extracted from the hyperspectral image to form an RGB image. The wavelengths are R = 650.45nm, G = 540.62nm, B = 460.27nm. The RGB image is then passed through the trained YOLO model, which detects the tomatoes and assigns a mask and an appropriate bounding box. The bounding box is subsequently used to extract the ROIs from the hyperspectral cube.  Let's define the bounding box from the YOLO model as:

 \begin{equation}
  B_{rgb} = [X_o, Y_o, H_o, W_o ]
  \label{eq:bb}
\end{equation}

Where $X_o$ and $Y_o$ are the bottom left corner pixel coordinates of the bounding box. Also, $H_o$ and $W_o$ are the height and width of the bounding box. We define the dimension of the RGB image as ($H_{rgb}$, $W_{rgb}$) and that of the HSI as ($H_{hsi}$, $W_{hsi}$, $D_{hsi}$) where $H$, $W$ and $D$ are the height, width and depth respectively. Recall also that we are cropping in the spatial dimension, thus the spatial bounding box of the HSI can be obtained as follows:

 \begin{equation}
  B_{hsi} = B_{rgb} \times \begin{bmatrix}
\alpha_1 & 0 & 0 & 0 \\
0 & \alpha_2 & 0 & 0 \\
0 & 0 & \alpha_2 & 0 \\
0 & 0 & 0 & \alpha_1 \\
\end{bmatrix}
  \label{eq:bbhsi}
\end{equation}

where $\alpha_1$ and $\alpha_2$ are scaling factors defined as $\frac{W_{hsi}}{W_{rgb}}$ and $\frac{H_{hsi}}{H_{rgb}}$ respectively. However, since RGB image is extracted from the HSI and the YOLO output gives similar size, as such $\alpha_1 = \alpha_2 = 1$. 

Another pre-processing step involves using the foreground mask of the tomato obtained from the YOLO model to remove the background. This is accomplished by setting all intensity values across all wavelengths to 0 if the corresponding pixel value in the mask is 0, and maintaining the original intensity if the pixel value is 1. Mathematically, this can be expressed as

\begin{equation}
\mathcal{I}_{\text{final}}(x, y, \lambda) = 
\begin{cases} 
0 & \text{if } M(x, y) = 0 \\
\mathcal{I}(x, y, \lambda) & \text{if } M(x, y) = 1 
\end{cases}
\end{equation}

where \( \mathcal{I}(x, y, \lambda) \) represents the intensity at pixel \((x, y)\) for wavelength \(\lambda\), and \( M(x, y) \) represents the mask value at pixel \((x, y)\).

\subsection{Proposed Model Pipeline}
\subsubsection{Variational Autoencoder (VAE)}
The complete pipeline starts by having the input pre-processed as in Section \ref{subsec:process}. 
\begin{figure}[h!]
  \centering
  \includegraphics[height=5cm,width=11.5cm]{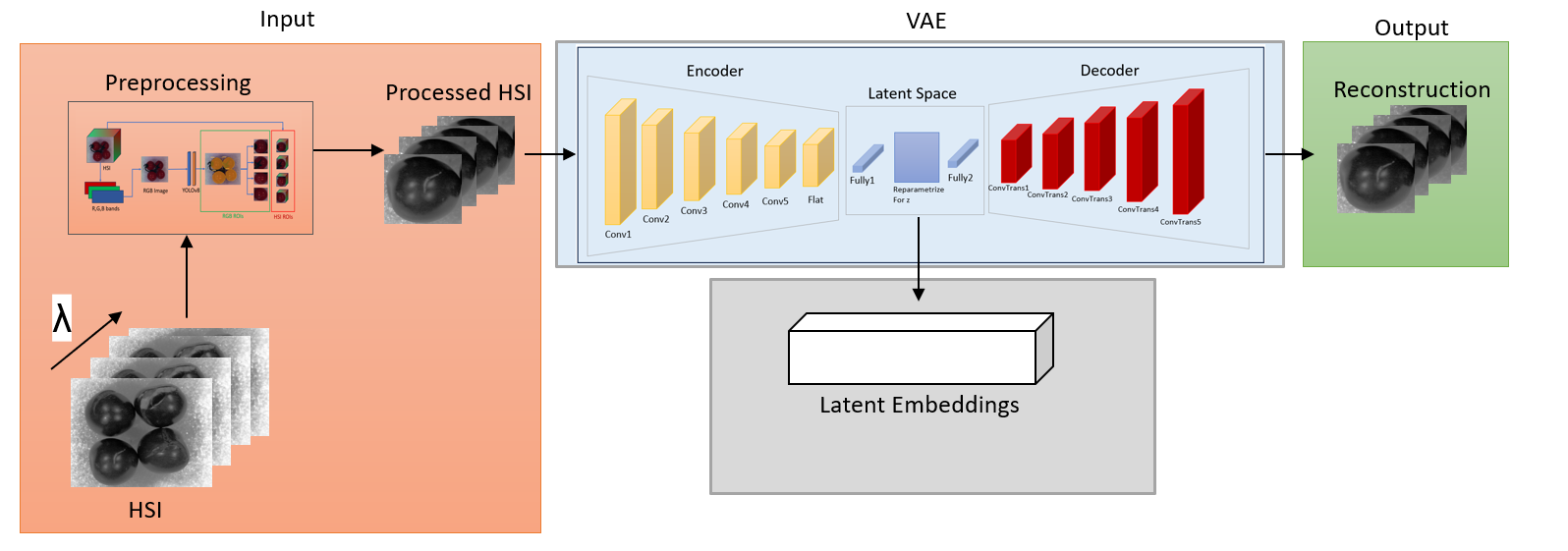}
  \caption{The proposed pipeline consisting of the input, VAE and the output. 
  }
  \label{fig:yolopreprocess}
\end{figure}
The processed input is passed to the VAE. As earlier discussed that our input ($x$) has a dimension ($H$, $W$, $D$) corresponding to Height, Width and Depth (spectral channels) of the HSI, the encoder consists of several convolutional layers compressing the input to give an output of two vectors: $\mu$ which is the mean and $V =log\sigma^2$ which is the logarithm of the variance of the latent space distribution. Suppose the encoding process is $E(x)$, we can say that: 
\begin{equation}
    \mu = f_\mu(E(x))
\end{equation}
and,
\begin{equation}
    V = f_V(E(x))
\end{equation}
where $f_\mu$ and $f_V$ are fully connected layers. 
The output of these functions $\mu$ and $V$ are then used to create latent variable $z$. To enable backpropagation, $z$ is sampled through a stochastic process. Since the direct sampling of $z$ from the normal distribution $\mathcal{N}(\mu, \sigma^2)$ may involve a non-differentiable operation which may hinder backpropagation, reparameterization trick is used to sample $z$ as follows:
firstly, a random variable $\epsilon$ is sampled from a standard normal distribution $\mathcal{N}(0, I)$.
\begin{equation}
    \epsilon \sim \mathcal{N}(0, I)
\end{equation}
where $I$ is an identity matrix. Next, we shift and scale $\epsilon$ using the $\mu$ and $\sigma$ from the encoder to obtain the sampled $z$ as follows. 
\begin{equation}
    z = \mu + \sigma \odot \epsilon
    \label{latent}
\end{equation}
$\sigma$ is also expressed as $\sigma = e^{(0.5 \cdot V)}$ to make the value positive, and $\odot$ denotes element wise multiplication. 

After obtaining the latent variable $z$, the decoder is then used to map the latent variable back to the hyperspectral data space, producing a reconstruction $\hat{x}$.
\begin{equation}
    \hat{x} = g(z)
\end{equation}

The loss function of the VAE combines a reconstruction loss and KL divergence loss. The reconstruction loss evaluates the accuracy of the decoder in reconstructing the hyperspectral image from its latent representation. To enhance sensitivity in detecting small anomalies, we utilize the L1 loss, which calculates the absolute difference between the reconstructed image and the original.  The reconstruction loss is defined as:
\begin{equation}
    \mathcal{L}_{\text{recon}} = \sum_{h=1}^{H} \sum_{w=1}^{W} \sum_{d=1}^{D} \left| x_{h,w,d} - \hat{x}_{h,w,} \right|
    \label{equ:reconloss}
\end{equation}

The second part of the loss function is the KL divergence loss which measures the divergence between the learned latent distribution $q(z|x)$ and the prior distribution $p(z)$ as follows:
\begin{equation}
    \mathcal{L}_{KL} = -\frac{1}{2} \sum_{i=1}^{S} \left( 1 + \log(\sigma_i^2) - \mu_i^2 - \sigma_i^2 \right)
\end{equation}
 where $S$ is the dimensionality of the latent space. 

Thus, we can now compute our loss function $\mathcal{L}_{VAE}$ as follows:
\begin{equation}
    \mathcal{L}_{VAE} = \mathcal{L}_{recon} + \beta \cdot \mathcal{L}_{KL}
\end{equation}

where $\beta$ is a weighting factor that controls the contribution of the losses. The model is trained to minimize $\mathcal{L}_{VAE}$ by back-propagation and updating the network parameters accordingly.  The overall VAE architecture is summarised below:

\begin{table}[t!]
  \caption{VAE architecture for tomato split detection (B is the batch size)
  }
  \label{tab:headings}
  \centering
  \begin{tabular}{@{}lll@{}}
    \toprule
    \textbf{Layer Type} & \textbf{Input Shape} & \textbf{Output Shape (activation)} \\
    \midrule
    \textbf{Encoder} & & \\
    \midrule
    Convolutional Layer & (B, 16, 210, 210) & (B, 16, 105, 105) (ReLU)\\
    Convolutional Layer & (B, 16, 105, 105) & (B, 32, 53, 53) (ReLU)\\
    Convolutional Layer & (B, 32, 53, 53) & (B, 64, 27, 27) (ReLU)\\
    Convolutional Layer &  (B, 64, 27, 27) & (B, 128, 14, 14) (ReLU)\\
    Convolutional Layer &  (B, 128, 14, 14) & (B, 256, 7, 7) (ReLU)\\
    \midrule
    \textbf{Latent Space} & & \\
    \midrule
    Fully Connected Layer & (B, 256 $\times$ 7 $\times$ 7) & (B, 100)\\
    Fully Connected Layer & (B, 256 $\times$ 7 $\times$ 7) & (B, 100)\\
    Reparameterization  & (B, 100), (B, 100) \ & (B, 100)\\
    Fully Connected Layer (decode) \ & (B, 100) & (B, 256 $\times$ 7 $\times$ 7)\\
     \midrule
    \textbf{Decoder} & & \\
    \midrule
    Transposed Convolutional Layer \ & (B, 256, 7, 7) & (B, 128, 14, 14) (ReLU)\\
    Transposed Convolutional Layer & (B, 128, 14, 14) &  (B, 64, 27, 27) (ReLU)\\
    Transposed Convolutional Layer& (B, 64, 27, 27) & (B, 32, 53, 53) (ReLU)\\
    Transposed Convolutional Layer& (B, 64, 27, 27) & (B, 32, 53, 53) (ReLU)\\
    Transposed Convolutional Layer & (B, 32, 53, 53) & (B, 16, 105, 105) (ReLU)\\
    Transposed Convolutional Layer & (B, 16, 105, 105) & (B, 16, 210, 210) (Sigmoid)\\
  \bottomrule
  \end{tabular}
\end{table}
\subsection{Training}
\begin{figure}[t!]
  \centering
  \includegraphics[height=4.5cm,width=7cm]{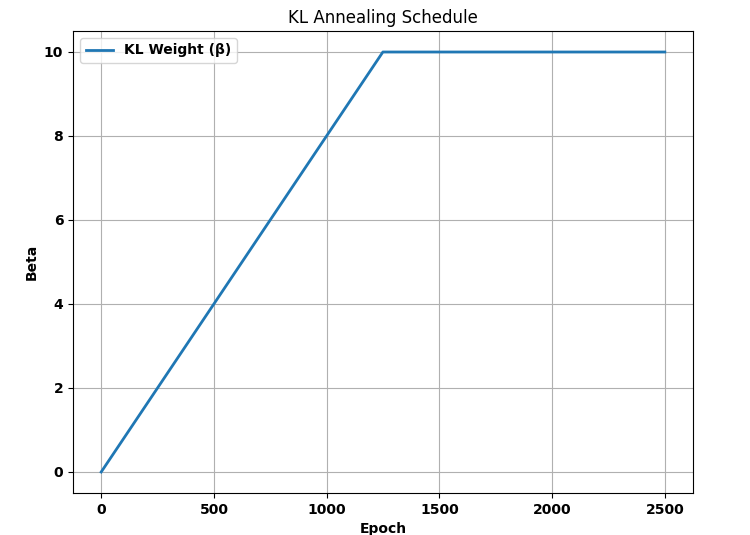}
  \caption{KL annealing schedule profile over the training span.
  }
  \label{fig:schedule}
\end{figure}

The network was trained using the processed data as described in Section \ref{subsec:process}. The model training spanned 2500 epochs, with the latent space dimensionality $S$ set to 100. Adam optimiser \cite{kingma2014adam} was used and the learning rate was set to $10^{-3}$ with a batch size of 32. A fixed $\beta$ value would cause the KL loss to dominate during the early stages of training, hindering the model from learning meaningful latent representations. To mitigate this, we employed the KL annealing technique, which gradually increases $\beta$, allowing the model to initially focus on reconstruction. The annealing schedule is defined to linearly increase during the first half of the training and remain constant thereafter, as shown in Equation \ref{eq:annealing}. Data is split 80\% training and 20\% testing.  As this is an unsupervised technique, the training dataset encompassed mainly the normal instances. While, the testing encompassed all the anomalous instances with few unseen normal instances. 
\begin{equation}
    \beta(t) = 
\begin{cases} 
\frac{2t}{T} \beta_{\text{max}} & \text{if } t \leq \frac{T}{2} \\
\beta_{\text{max}} & \text{if } t > \frac{T}{2}
\end{cases}
\label{eq:annealing}
\end{equation}
where $t$ is the current epoch, $\beta_{\text{max}}$ is the maximum weight set at 10, $T$ is the total number of epoch set at 2500. The annealing schedule can be visualised in Fig. \ref{fig:schedule}.
\section{Results}
\label{sec:result}

\subsection{Wavelengths of Interest}
The initial experiment aimed to identify the wavelengths of interest to enhance the accuracy of our model and reduce computational load by pinpointing key wavelengths while discarding redundant ones. We approached this task through both visual analysis and inspection of the reflectance.

\begin{figure}[h!]
  \centering
  \includegraphics[height=4.5cm,width=11.8cm]{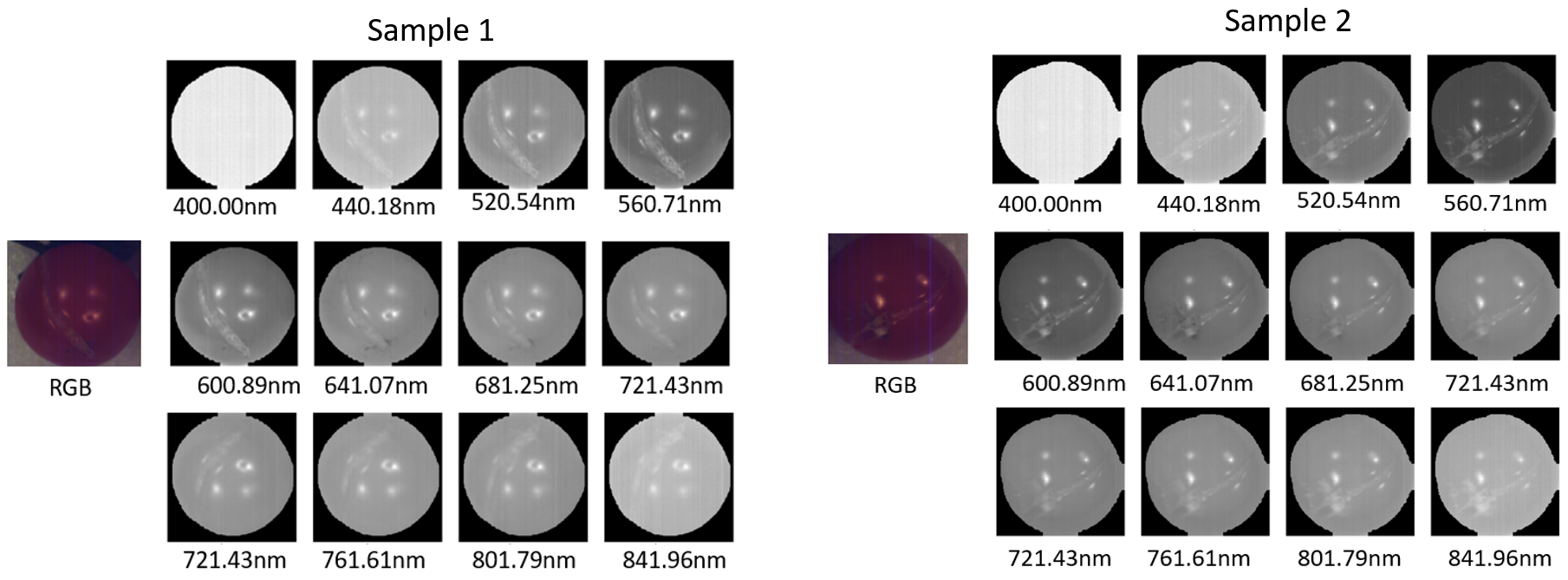}
  \caption{Visualisation of tomato split samples across several wavelengths.
  }
  \label{fig:visualsplit}
\end{figure}

In the visual analysis, we selected a range of wavelengths that encompass the entire spectrum and manually inspected the images to determine where the tomato split was most apparent. For example, in Figure \ref{fig:visualsplit}, two sample images demonstrate that the split is most apparent in the wavelength range of 520.54 nm to 600.89 nm. 

\begin{figure}[t!]
  \centering
  \includegraphics[height=3.5cm,width=11.8cm]{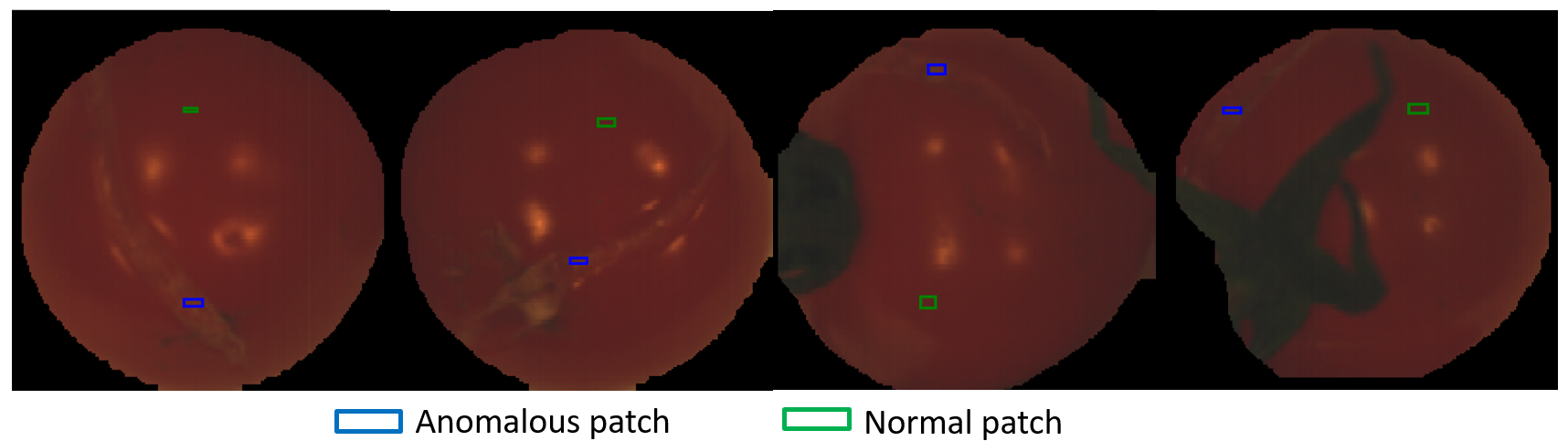}
  \caption{Patch selection from both normal and anomalous region taken from 4 different samples. 
  }
  \label{fig:splitpatch}
\end{figure}

For the reflectance analysis, we selected two patches: one from the split region and another from a normal region and compared their spectral responses. As shown in Figure \ref{fig:splitpatch}, the patches (illustrated in RGB) were selected across four different HSI images. We then computed the average reflectance of the pixels within each patch. By analyzing these average reflectance values, we could identify the specific wavelengths that exhibited significant differences between the split and normal regions. 

\begin{figure}[h]
  \centering
  \includegraphics[height=7cm,width=11cm]{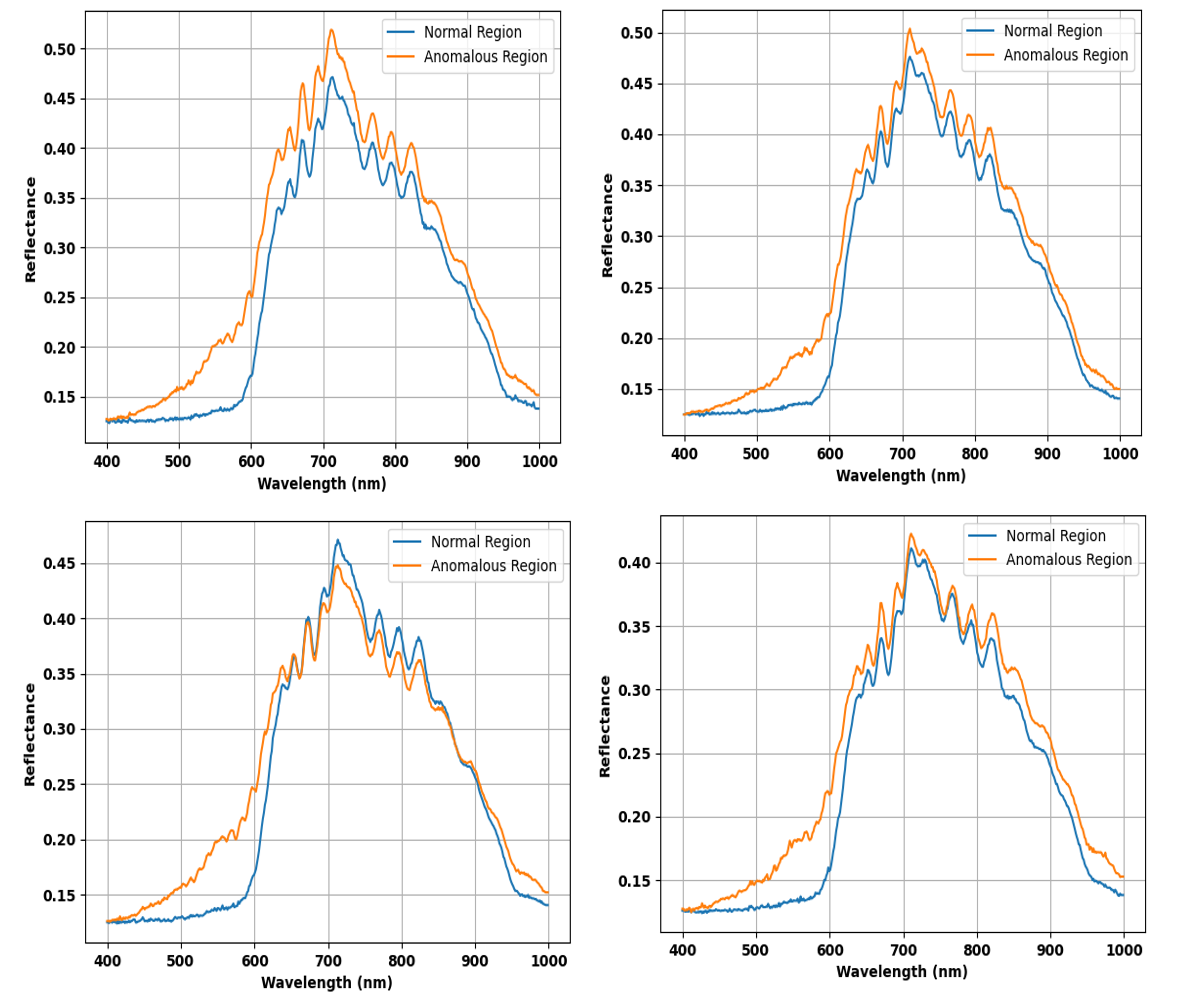}
  \caption{Corresponding average reflectance for the selected patches across the 4 different samples. The difference in reflectance for normal and anomalous regions can be observed in the 520nm - 600nm range.
  }
  \label{fig:refelectancevalue}
\end{figure}

From Figure \ref{fig:refelectancevalue}, we can observe that the reflectance difference between the normal patch and the anomalous patch is more pronounced within our previously identified range, confirming the findings from the visual inspection. Additionally, this range is consistently significant across the four selected frames, reinforcing the reliability of this selection. Thus, we selected the range 530nm to 550nm for the experiments.  

\subsection{VAE Results}
\begin{figure}[t!]
  \centering
  \includegraphics[height=5cm,width=12cm]{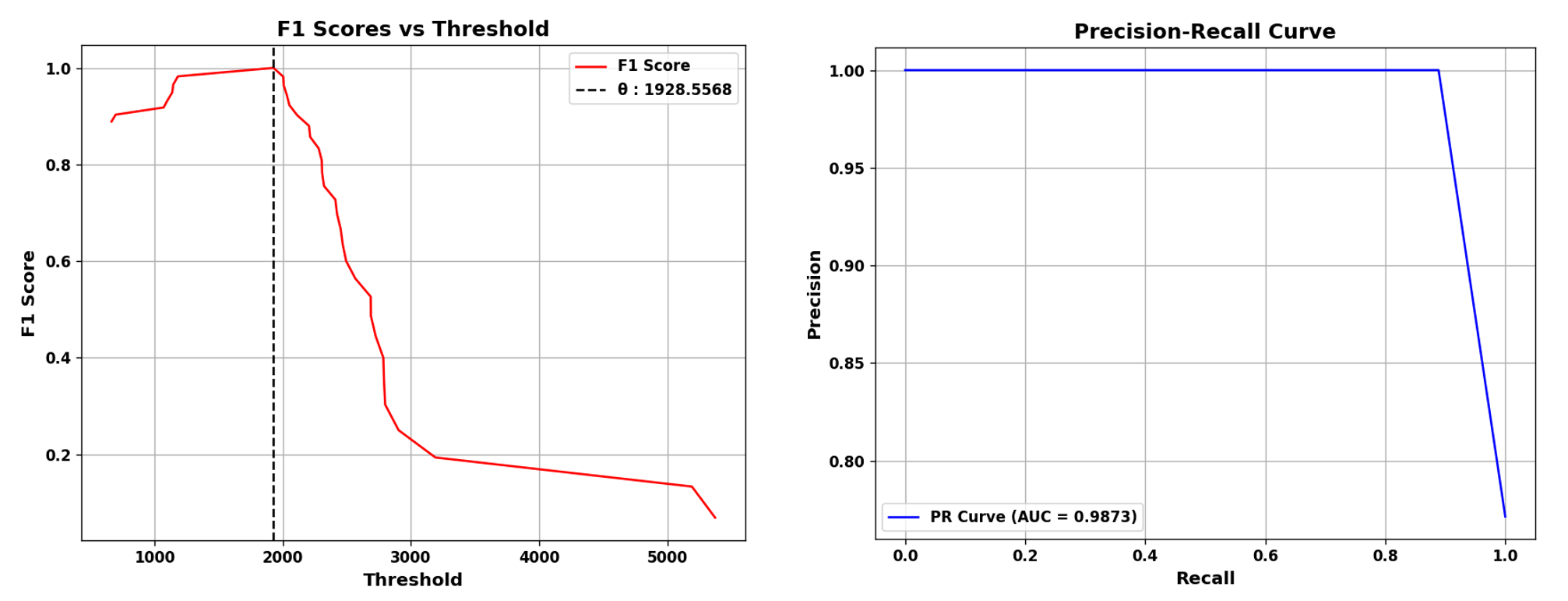}
  \caption{F1 score across a set of thresholds to obtain the optimal threshold. The Precision and Recall curve for the unseen set using the obtained threshold. 
  }
  \label{fig:precisionrecall}
\end{figure}

The reconstruction loss, derived from Equation \ref{equ:reconloss} was employed to determine the anomaly status of the tomato. The objective is to establish a threshold that effectively distinguishes between normal and anomalous tomatoes. The threshold is determined by splitting the testing dataset, which comprises both normal and anomalous instances, into two halves to mitigate bias. Precision and recall were computed for the first half, yielding the F-score. The optimal threshold was estimated by evaluating the F-score across thresholds and selecting the threshold that maximized the score, as depicted in Figure \ref{fig:precisionrecall}. Subsequently, we validated this selected threshold on the remaining unseen data, generating the precision-recall plot shown in Figure \ref{fig:precisionrecall}. For different threshold $\theta$ in a set of thresholds $\Theta$ depicting the range of reconstruction losses, the maximum $F1$ can be computed as:

\begin{equation}
\text{F1}_{\text{max}} = \max_{\theta \in \Theta} \left( \frac{2 \cdot \text{Precision}(\theta) \cdot \text{Recall}(\theta)}{\text{Precision}(\theta) + \text{Recall}(\theta)} \right)
\end{equation}

Since the model was trained exclusively on normal data, the threshold is such that any reconstruction error exceeding this value indicates an anomaly. Conversely, errors below this threshold denote normal tomato. The threshold criterion for anomaly detection is defined by the following equation:

\begin{equation} 
\text{Status} =
\begin{cases} 
\text{Anomalous} & \text{if} \ \mathcal{L}_{\text{recon}} > \theta \\
\text{Normal} & \text{otherwise}
\end{cases}
\end{equation}

\begin{figure}[t!]
  \centering
  \includegraphics[height=5cm,width=12cm]{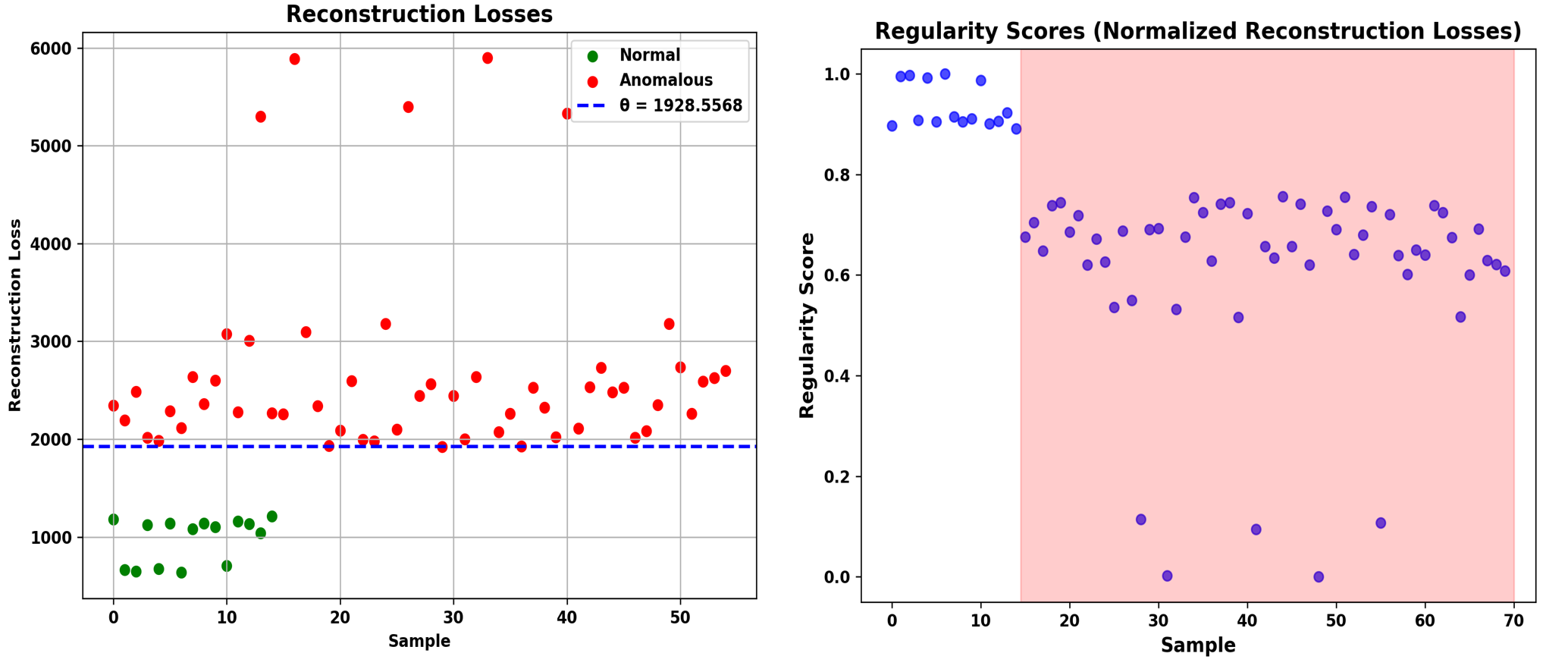}
  \caption{Reconstruction loss and regularity score for the test data containing both normal and anomalous samples.
  }
  \label{fig:anomresult}
\end{figure}

Figure \ref{fig:anomresult} shows the reconstruction losses and the regularity scores of the test data, including both normal and anomalous instances. The regularity score is the normalised reconstruction loss. We have a total of 15 normal instances and 55 anomalous instances in the test set. The threshold ($\theta$) $ = 1928.6$ previously obtained effectively distinguishes between normal and anomalous tomato. From the figure, it can be observed that the reconstruction errors for normal instances are below the threshold, while the errors for anomalous instances are above the threshold. Although two of the anomalous instances are slightly below the threshold, the clear separation demonstrates the efficacy of the model in identifying tomato split based on reconstruction error.

\begin{figure}[t!]
  \centering
  \includegraphics[height=5.5cm,width=12cm]{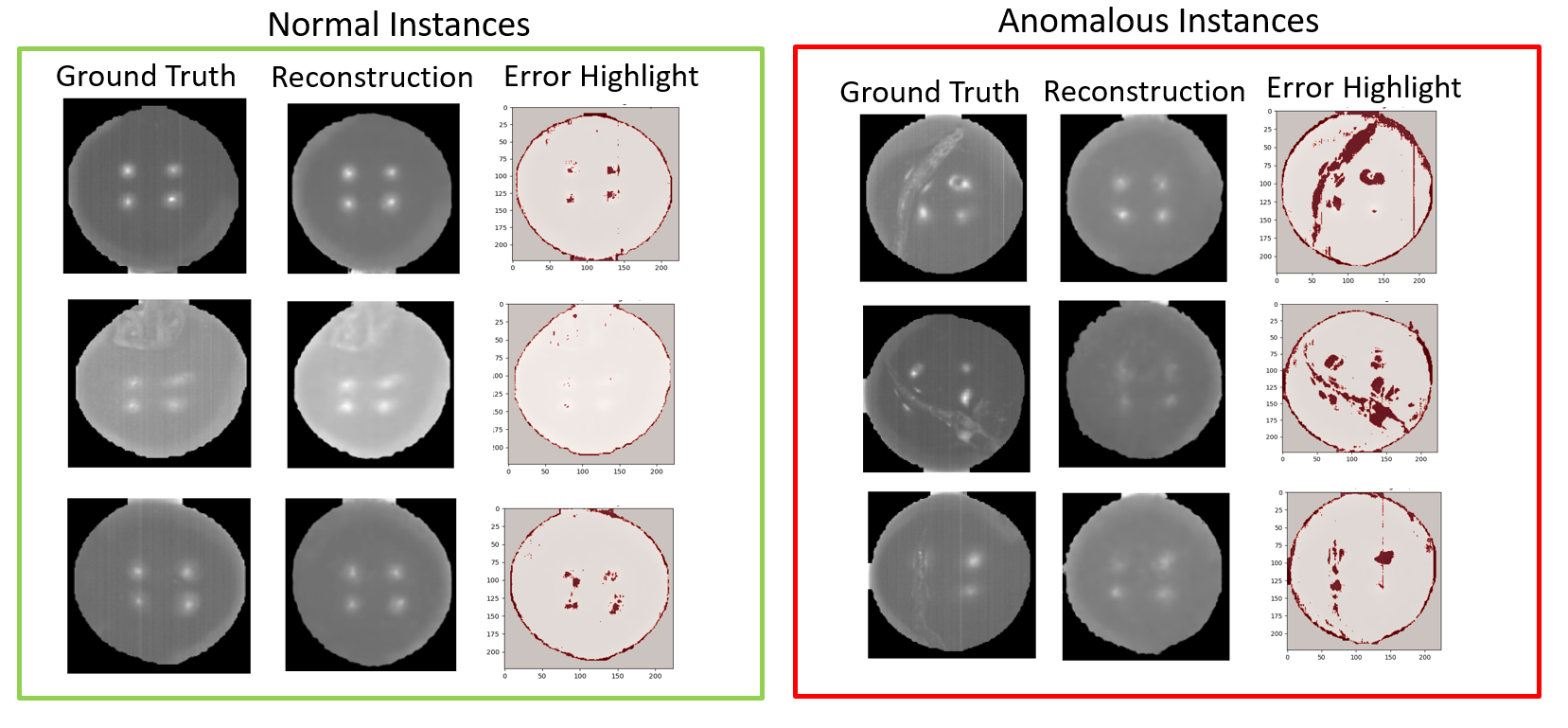}
  \caption{Qualitative results from both normal and anomalous samples. The split region is highlighted using a mask.
  }
  \label{fig:qualitative}
\end{figure}

 Figure \ref{fig:qualitative}  shows some qualitative results of the VAE for samples from both normal and anomalous instances. We compare the ground truth and the reconstruction, as well as add a mask highlighting the contribution of the pixels to the overall reconstruction loss based on a threshold. It is observed that the split region significantly contributes to the overall reconstruction loss in the anomalous instances. Conversely, in the normal instances, the reconstruction error is relatively minimal, indicating the absence of the split. Thus, to some extent, the model is also capable of highlighting the split regions.

\begin{figure}[t!]
  \centering
  \includegraphics[height=5cm,width=11cm]{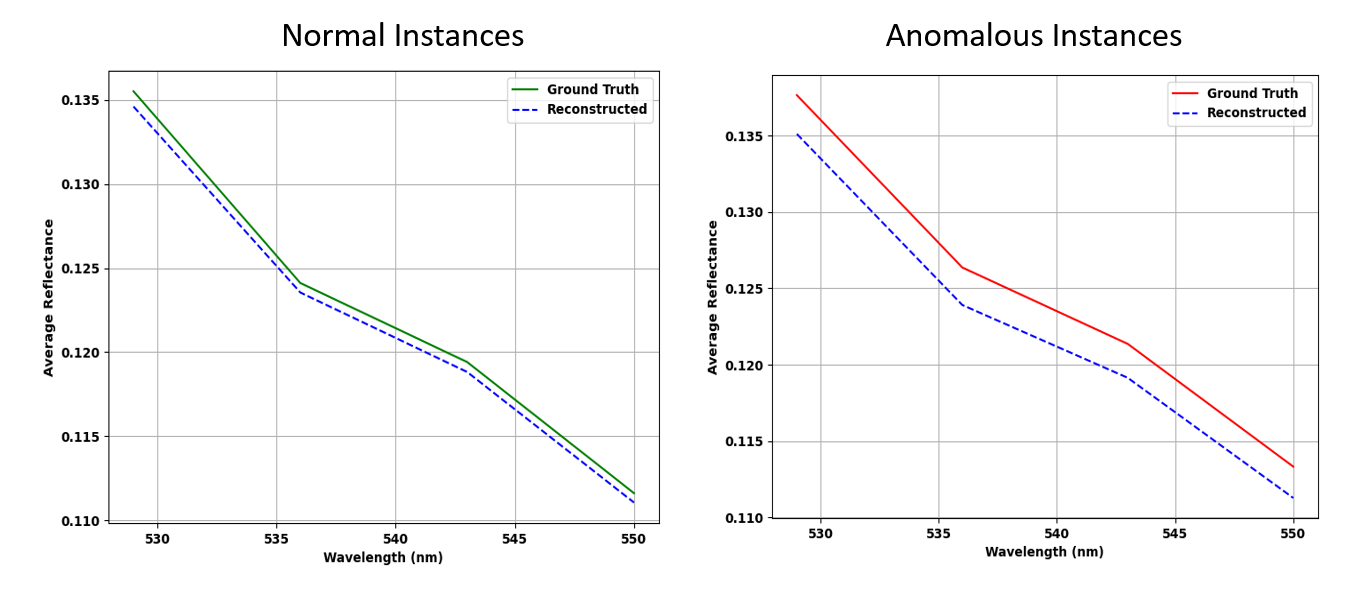}
  \caption{Ground truth and reconstructed average reflectance for both normal and anomalous instances of the test dataset. Green is the normal instance ground truth, red is the anomalous instance ground truth.
  }
  \label{fig:avgreflectanceall}
\end{figure}

Finally, we conducted a comparative analysis on the average reflectance across the entire dataset for both normal and anomalous instances in Fig. \ref{fig:avgreflectanceall}. This involved calculating the mean reflectance at each wavelength for all samples in the test dataset. The ground truth is the reflectance obtained from the original HSI which we then compare with the model's reconstruction. From this analysis, it can be observed that the average reflectance for the normal data closely aligns with the ground truth reflectance values. In contrast, the anomalous data exhibits a less precise estimation, indicating a higher reconstruction error.
\section{Conclusion}
\label{sec:conc}
In this study, we have presented an effective approach for detecting tomato split anomalies using a tailored variational autoencoder (VAE) with hyperspectral input. Our preliminary analysis identified the optimal wavelength range of 530nm - 550nm for detecting tomato split anomalies. The reconstruction loss of the proposed VAE was used as a metric to identify splits by selecting a suitable threshold. The VAE model demonstrated high detection accuracy by effectively having a lower reconstruction loss for the normal samples and relatively higher loss for the anomalous samples thus highlighting its potential for improving quality control in greenhouse farming.

Future work may focus on adding a classification head to classify the extent of the splits for grading purposes. Additionally, exploring anomaly detection with sufficient dataset and without the specularity of light on the target could further increase the accuracy of the model.

\bibliographystyle{splncs04}
\bibliography{main}
\end{document}